\documentclass[runningheads]{llncs}

\usepackage{cite}
\usepackage{amsmath,amssymb,amsfonts}
\usepackage{algorithm}
\usepackage{algorithmic}
\usepackage{tikz}
\usetikzlibrary{positioning, arrows.meta, shapes}
\usepackage{subcaption}
\usepackage{graphicx}
\usepackage{textcomp}
\usepackage{xcolor}
\usepackage{braket}

\begin{document}

\title{Quantum-Evolutionary Neural Networks for Multi-Agent Federated Learning}

\author{
  Aarav Lala\inst{1*} \and 
  Kalyan Cherukuri\inst{1}\thanks{Equal contribution by Aarav Lala and Kalyan Cherukuri.}
}
\institute{
  Institution Name \\
  \email{\{aarav.lala,kalyan.cherukuri\}@example.com}
}

\authorrunning{A. Lala and K. Cherukuri}

\institute{Illinois Mathematics and Science Academy, Aurora, IL, USA\\
\email{\{alala1,kcherukuri\}@imsa.edu}}

\maketitle

\begin{abstract}
As artificial intelligence continues to drive innovation in complex, decentralized environments, the need for scalable, adaptive, and privacy-preserving decision-making systems has become critical. This paper introduces a novel framework combining quantum-inspired neural networks with evolutionary algorithms to optimize real-time decision-making in multi-agent systems (MAS). The proposed Quantum-Evolutionary Neural Network (QE-NN) leverages quantum computing principles—such as quantum superposition and entanglement—to enhance learning speed and decision accuracy, while integrating evolutionary optimization to continually refine agent behaviors in dynamic, uncertain environments. By utilizing federated learning, QE-NN ensures privacy preservation, enabling decentralized agents to collaborate without sharing sensitive data. The framework is designed to allow agents to adapt in real-time to their environments, optimizing decision-making processes for applications in areas such as autonomous systems, smart cities, and healthcare. This research represents a breakthrough in merging quantum computing, evolutionary optimization, and privacy-preserving techniques to solve complex problems in multi-agent decision-making systems, pushing the boundaries of AI in real-world, privacy-sensitive applications.
\end{abstract}

\section{Introduction}

In recent years, artificial intelligence has experienced remarkable forward progress, especially as it regards to this study within the realm of decentralized decision-making systems. This systems, commonly referred to as multi-agent systems (MAS), consist of independent agents interacting amongst one another within a complex environment. This recent push has made scalable, adaptive, and privacy-preserving solutions in MAS a pressing need \cite{jenefa2024enhancing, yun2023quantum}.

A notable challenge towards development of these robust MAS is the ability to ensure accurate real-time decisions while maintaining privacy across distributed agents. Traditional centralized approaches still fall short in these scenarios, requiring the pooling of sensitive data across multiple agents, a growing privacy concern for real-world implementations. Currently, static decision-making models are unable to adapt to the dynamic and uncertainty present in the environment that these agents operate in. Our approach proposes integration of quantum computing into existing techniques to enhance the speed, efficiency, and adaptability of these decision-making processes \cite{ortiz2023quantum, friedrich2023evolution}.

This study presents a novel framework, Quantum-Evolutionary Neural Networks, leveraging popular quantum computing principles of superposition and entanglement to improve learning efficiency while mainting similar results to baselines. Furthermore, our proposed evolutionary algorithm in coordination with federated learning enables real-time agent evolution while still preserving privacy of individual data \cite{cagnoni2023federated}. This research aims to push the existing boundaries of artificial intelligence through a combination of quantum computing, evolutionary optimization, and privacy-preserving techniques for multi-agent decision-making.

\section{Related Works}

\subsection{Federated Learning under Non-IID Data Distributions}

Federated Learning enables decentralized privacy-preserving model training, however; it faces some challenges. These challenges arise specifically with non-independent and identically distributed (non-IID) data across clients. This data impairs model convergence and generalization ability across clients, fortunately extensive literate has studied this presenting architectural, algorithmic, and data-centric solutions.

A theoretical framework developed to better understand the impact of data heterogeneity for adaptive algorithms improving convergence \cite{wei2021federated}. This work laid the foundation for more robust algorithm design in non-IID settings.

One of the earliest and most influential works addressing non-IID data in FL introduced FedAvg, a simple yet effective aggregation strategy that revealed the extent to which data heterogeneity could hinder convergence \cite{mcmahan2017communication}. While not originally designed to solve non-IID challenges, FedAvg highlighted the limitations of naive averaging and motivated the development of more sophisticated solutions tailored to statistical heterogeneity.

In 2019, a new framework was introduced, Sparse Ternary Compression(STC)\cite{sattler2019robust}. This method reduced communication cost using sparsification and ternarization of gradients, but preserved the model's robustness in heterogenous data environments. STC demonstrated that compression and accuracy
does not need to be mutually exclusive, especially under non-IID distributions.

Another algorithm that was proposed was FedDC, an algorithm that prioritizes decouples and corrects local model drift-one of the key symptoms of non-IID training \cite{gao2022feddc}. By introducing a proximal correction mechanism that aligns local updates with the global model’s trajectory, FedDC significantly improves convergence rates and final model accuracy, even when clients have severely skewed data distributions.

A major issue with federated learning is its vulnerability to client dropout and data heterogeneity, which can lead to biased model updates and degraded performance. This challenge was tackled thorough the usage of secure secret-sharing protocols\cite{shao2022coded}. By encoding shared data into coded pieces using Lagrange polynomial interpolation, their method ensured that the global gradient stayed unbiased, even when the clients drop out or act maliciously.

Together, these works demonstrate a broad spectrum of solutions to the non-IID problem—ranging from data augmentation and gradient compression to drift correction and coded computation. Despite their diversity, a common theme unites them: the critical need to reconcile decentralized training objectives with a coherent global model, especially in the presence of heterogeneity that reflects real-world applications.

\subsection{Evolutionary Algorithms in Distributed Optimization}
Evolutionary algorithms (EAs) are an emerging and robust alternative to gradient-based optimization in distributed settings \cite{chai2023communication}, particularly when facing non-differential non-convex, or noisy objective landscapes. Inspired through natural selection, EAs operate through candidate solutions, applying operations from mutation, crossover, and selections to iteratively evolve towards optimal solutions. 
In distributed optimization, especially in federated learning (FL) tasts, EAs present unique advantage to traditional gradient-based methods that suffer from issues ranging from data heterogeneity, communication bottlenecks, and non-convexity. Evolutionary strategies, being free of the gradient, are more resilient to such issues and can handle issues prior like partial participation, asynchronous updates, and even adversarial agents more gracefully \cite{morell2022optimising}.
Several works, have demonstrated the exploration between EAs and distributed optimization \cite{pmlr-v238-zakerinia24a}. For example, consider the proposed Population Based Training (PBT) \cite{jaderberg2017population}, where evolutionary concepts were demonstrated for optimizing hyperparameters and models in parallel environments. Similarly, other strategies such as neuroevolution for deep learning in environments characterized by high distributions bypass the purpose for standard backpropogation all together. \cite{miikkulainen2019evolving}
In a federated environment, there have been past mechanisms such as FedSel, an evolutionary selection mechanism which identies high-quality local models to aggregate, improving robustness and performance while iterating through clients \cite{liu2020fedsel}. EvoFed \cite{rahimi2023evofed} extends this idea through leveraging of evolutionary strategies for communication overhead reductions in FL systems, presenting a scalable and gradient-free alternative for model updates. Recently, hybrid methods combining evolutionary search with gradient-based updating for faster convergence has demonstrated enhanced exploration-exploitation tradeoffs in FL \cite{yu2023evolutionary}.

\subsection{Quantum-Inspired Neural Architectures and Privacy Mechanisms}
The integration of quantum computing principles into neural network architectures has led to the development of quantum-inspired neural networks, which aim to leverage quantum phenomena to enhance computational capabilities.
By utilizing continuous degrees of freedom such as the amplitudes of electromagnetic fields, continuous variable quantum neural networks have been proposed as the main framework for combining quantum machine learning with layered structures similar to classical neural networks\cite{killoran2019continuous}. 
To address the many challenges of nonlinear activation in quantum systems, nonlinear quantum neurons have been introduced as a fundamental unit for quantum neural networks. These neurons work by integrating measurement based operations to emulate nonlinearity, allowing for more expressive models within a quantum regime\cite{yan2020nonlinear}. 
Another major advancement is the development of coherent feed-forward quantum neural networks, which use coherent quantum operations to mimic the feed-forward behavior in class networks. This approach has been shown to maintain the advantages of quantum coherence while enabling modular construction of deeper architectures \cite{singh2024coherent}.
Beyond architecture, neural networks have been employed to solve quantum dynamical problems. For instance, artificial neural networks have been used as propagators to accelerate quantum simulations by learning to approximate time-dependent solutions to the Schrödinger equation.\cite{secor2021artificial}.

\section{Problem Formulation}
Consider a decentralized multi-agent system comprising of $N$ clients, where each hold a dataset locally $D_i = \{(x_j^{(i)}, y_j^{(i)})\}_{j=1}^{n_i}$, sampled from a non-identical and non-independent distribution $\mathcal{D}_i$. The goal is to train a global model $f_\theta: \mathbb{R}^d \to \mathbb{R}^c$, parameterized by $\theta \in \mathbb{R}^p$, to minimize empirical risk across all clients, without requiring access to raw data.

\subsection{Federated Learning Objective}

The standard federated learning objective is given by:

\begin{equation}
\min_{\theta \in \mathbb{R}^p} \; \mathcal{L}(\theta) = \sum_{i=1}^N \frac{n_i}{n} \cdot \mathcal{L}_i(\theta),
\end{equation}

where $n = \sum_{i=1}^N n_i$ is the total number of training examples and the local loss for client $i$ is:

\begin{equation}
\mathcal{L}_i(\theta) = \frac{1}{n_i} \sum_{j=1}^{n_i} \ell(f_\theta(x_j^{(i)}), y_j^{(i)}),
\end{equation}

with $\ell: \mathbb{R}^c \times \mathbb{R}^c \to \mathbb{R}$ denoting a loss function such as cross-entropy.

\subsection{Quantum-Evolutionary Neural Network (QE-NN)}

The model $f_\theta$ is instantiated as a Quantum-Evolutionary Neural Network (QE-NN), whose core layers are designed to simulate quantum interference via phase-shifted sine activations:

\begin{equation}
z^{(l)} = \sin(W^{(l)} z^{(l-1)} + \phi^{(l)}),
\end{equation}

where $W^{(l)} \in \mathbb{R}^{d_l \times d_{l-1}}$ are learnable weights and $\phi^{(l)} \in \mathbb{R}^{d_l}$ are trainable phase shift parameters.

This periodic architecture enables the network to capture complex, nonlinear interactions between features, inspired by quantum superposition and entanglement.

\subsection{Local Evolutionary Optimization}

Each client $i$ generates $K$ locally mutated variants of the global model by applying Gaussian perturbations:

\begin{equation}
\theta_i^{(k)} = \theta + \epsilon_k^{(i)}, \quad \epsilon_k^{(i)} \sim \mathcal{N}(0, \sigma^2 I), \quad k = 1, \dots, K
\end{equation}

Each variant is locally fine-tuned, and the client selects the best-performing variant:

\begin{equation}
\theta_i^\star = \arg\min_{\theta_i^{(k)}} \mathcal{L}_i(\theta_i^{(k)})
\end{equation}

\subsection{Privacy-Preserving Aggregation}

To ensure privacy, noise is added to each client’s selected model before transmission:

\begin{equation}
\tilde{\theta}_i = \theta_i^\star + \delta_i, \quad \delta_i \sim \mathcal{N}(0, \sigma_p^2 I)
\end{equation}

The server aggregates the models using federated averaging:

\begin{equation}
\theta \leftarrow \frac{1}{N} \sum_{i=1}^N \tilde{\theta}_i
\end{equation}

\subsection{Final Optimization Objective}

The overall procedure approximates the solution to the following privacy-regularized stochastic optimization problem:

\begin{equation}
\min_{\theta \in \mathbb{R}^p} \; \sum_{i=1}^N \frac{n_i}{n} \cdot \mathbb{E}_{\epsilon_k^{(i)}, \delta_i} \left[ \min_{k \in \{1, \dots, K\}} \mathcal{L}_i(\theta + \epsilon_k^{(i)}) + \frac{\lambda}{2} \|\delta_i\|^2 \right]
\end{equation}

subject to the constraint that raw data $D_i$ is never shared across clients or with the server. The inner minimization represents the evolutionary selection mechanism, and the expectation captures both mutation and privacy-preserving noise. The regularization term $\frac{\lambda}{2} \|\delta_i\|^2$ enforces a privacy budget, analogous to differential privacy frameworks.

\section{Theoretical Validation}

We provide theoretical justification for the convergence and privacy guarantees of the proposed Quantum-Evolutionary Federated Learning framework.

\subsection{Proof of Convergence}

Let $\mathcal{L}(\theta)$ denote the global loss function as previously defined, with these conditions holding:

\begin{enumerate}
    \item \textbf{Smoothness:} Each local loss function $\mathcal{L}_i(\theta)$ is $L$-smooth, i.e.,
    \[
    \|\nabla \mathcal{L}_i(\theta) - \nabla \mathcal{L}_i(\theta')\| \leq L \|\theta - \theta'\|, \quad \forall \theta, \theta' \in \mathbb{R}^p.
    \]
    
    \item \textbf{Bounded Gradient Norm:} The gradients of local losses are uniformly bounded, i.e.,
    \[
    \|\nabla \mathcal{L}_i(\theta)\| \leq G, \quad \forall i \in \{1, \dots, N\}, \; \theta \in \mathbb{R}^p.
    \]
    
    \item \textbf{Bounded Perturbation Variance:} The perturbation and privacy noises are zero-mean and have bounded variance:
    \[
    \mathbb{E}[\|\epsilon_k^{(i)}\|^2] \leq \sigma^2, \quad \mathbb{E}[\|\delta_i\|^2] \leq \sigma_p^2.
    \]
\end{enumerate}
Let $\theta^t$ denote the model at global round $t$. Then under assumptions (1)–(3), the expected reduction in loss over one round satisfies:
\begin{equation}
\mathbb{E}\left[\mathcal{L}(\theta^{t+1})\right] \leq \mathcal{L}(\theta^t) - \eta \|\nabla \mathcal{L}(\theta^t)\|^2 + \eta^2 C(\sigma^2 + \sigma_p^2),
\end{equation}
where $\eta$ is the effective learning rate induced by local evolution and $C$ is a constant dependent on $L$ and $G$.

This implies that as $t \to \infty$ and with properly tuned mutation and noise variances, the model converges to a stationary point of $\mathcal{L}(\theta)$ in expectation:

\begin{equation}
\lim_{T \to \infty} \frac{1}{T} \sum_{t=1}^T \mathbb{E} \left[ \|\nabla \mathcal{L}(\theta^t)\|^2 \right] \to 0.
\end{equation}

\subsection{Superposition and Entanglement}

Our model leverages quantum-inspired principles—specifically superposition and entanglement—to enhance the expressive capacity and diversity of local client models in federated learning. In classical quantum mechanics, a system in superposition can exist in multiple states simultaneously. We simulate this concept using the custom \texttt{QuantumLayer}, where input activations are transformed via linear projection and passed through a sinusoidal activation function combined with a learnable phase shift:
\[
x \mapsto \sin(Wx + \phi),
\]
where \( \phi \in \mathbb{R}^d \) is a trainable phase vector, simulating interference effects. This sinusoidal activation introduces periodicity and nonlinearity that helps the model capture richer representational dynamics, similar to how quantum states encode multiple possibilities simultaneously.

Quantum entanglement represents a coupling between quantum states such that the state of one cannot be described independently of the others. We emulate this phenomenon through stacked \texttt{QuantumLayer} blocks in the \texttt{QENN} architecture. Each layer’s output depends not only on the input features but also on the globally learned phase shift, effectively creating a shared encoding across the model's internal layers. This can be viewed as a form of functional entanglement, where transformations across layers are jointly conditioned and coupled.

\begin{figure}[t!]
    \centering
    \includegraphics[width=0.7\textwidth]{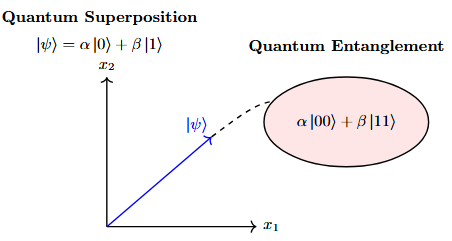}
    \caption{From superposition to entanglement: A single qubit $\ket{\psi}$ can be extended into an entangled two-qubit state $\alpha \ket{00} + \beta \ket{11}$.}
    \label{fig:loss}
\end{figure}

\subsection{Evolutionary Benefit \& Privacy Guarantee}

The inner evolutionary mechanism improves local optimization by exploring a richer landscape of parameter perturbations. Assuming Gaussian perturbations are sufficiently expressive to explore local minima, the probability of selecting a strictly better variant than \( \theta \) is strictly positive unless \( \theta \) is already a local minimum:
\begin{equation}
\mathbb{P}\left[\exists k \text{ s.t. } \mathcal{L}_i(\theta + \epsilon_k^{(i)}) < \mathcal{L}_i(\theta)\right] > 0 \quad \text{if } \nabla \mathcal{L}_i(\theta) \neq 0
\end{equation}
Thus, the evolutionary mechanism acts as a local improvement oracle, increasing the robustness of optimization, especially under non-convex loss landscapes common in neural networks. The addition of Gaussian noise \( \delta_i \sim \mathcal{N}(0, \sigma_p^2 I) \) ensures \( (\epsilon, \delta) \)-differential privacy under the Gaussian mechanism, given that the sensitivity \( \Delta \) of the update is bounded. For each round, if the sensitivity of model updates is bounded by \( \Delta \), then the per-round update satisfies:

\begin{equation}
(\epsilon, \delta)\text{-DP, where } \epsilon = \frac{\Delta^2}{2 \sigma_p^2}, \quad \delta \ll 1
\end{equation}

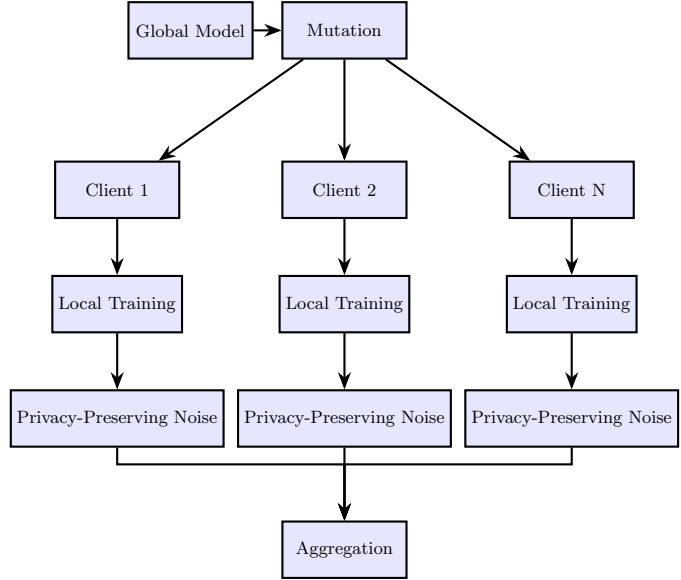
\begin{figure}[t!]
\centering
\begin{tikzpicture}[
    scale=0.75, 
    transform shape,
    font=\footnotesize, 
    node distance=1.0cm and 0.5cm, 
    every node/.style={align=center},
    box/.style={rectangle, draw=black, thick, minimum height=1cm, minimum width=2.2cm, fill=blue!10}, 
    arrow/.style={thick,->,>=Stealth}
]

\node[box] (global) {Global Model};
\node[box, right=of global] (mutation) {Mutation};

\node[box, below left=1.8cm and 1.8cm of mutation] (client1) {Client 1};
\node[box, below=of client1] (train1) {Local Training};
\node[box, below=of train1] (noise1) {Privacy-Preserving Noise};

\node[box, below=1.8cm of mutation] (client2) {Client 2};
\node[box, below=of client2] (train2) {Local Training};
\node[box, below=of train2] (noise2) {Privacy-Preserving Noise};

\node[box, below right=1.8cm and 1.8cm of mutation] (clientN) {Client N};
\node[box, below=of clientN] (trainN) {Local Training};
\node[box, below=of trainN] (noiseN) {Privacy-Preserving Noise};

\draw[arrow] (global) -- (mutation);
\draw[arrow] (mutation) -- (client1);
\draw[arrow] (mutation) -- (client2);
\draw[arrow] (mutation) -- (clientN);

\draw[arrow] (client1) -- (train1);
\draw[arrow] (train1) -- (noise1);

\draw[arrow] (client2) -- (train2);
\draw[arrow] (train2) -- (noise2);

\draw[arrow] (clientN) -- (trainN);
\draw[arrow] (trainN) -- (noiseN);

\node[box, below=1.3cm of noise2] (agg) {Aggregation};

\draw[arrow] (noise1.south) -- ++(0,-0.3) -| (agg.north);
\draw[arrow] (noise2) -- (agg);
\draw[arrow] (noiseN.south) -- ++(0,-0.3) -| (agg.north);

\end{tikzpicture}
\caption{Quantum-Evolutionary Federated Learning Pipeline}
\end{figure}

\subsection{Proposed Framework}
\vspace{-0.5 cm}

\begin{algorithm}[H]
\caption{Quantum-Evolutionary Federated Learning (QE-FL)}
\begin{algorithmic}[1]
\REQUIRE Initial global model $f_\theta$, number of clients $N$, local epochs $E$, learning rate $\eta$, mutation std $\sigma$, noise std $\sigma_p$, variants per client $K$, rounds $R$
\ENSURE Trained global model $f_\theta$
\FOR{$r = 1$ to $R$}
    \STATE Initialize empty list $\mathcal{M} \leftarrow []$ \hfill
    \FOR{each client $i = 1$ to $N$ \textbf{in parallel}}
        \STATE $(X_i, y_i) \leftarrow$ local dataset of client $i$
        \STATE Initialize $f^\star \leftarrow \text{None}, \; \mathcal{L}^\star \leftarrow \infty$
        \FOR{$k = 1$ to $K$}
            \STATE $f^{(k)} \leftarrow f_\theta + \epsilon_k$, \quad $\epsilon_k \sim \mathcal{N}(0, \sigma^2 I)$
            \FOR{$e = 1$ to $E$}
                \STATE Perform SGD update on $f^{(k)}$ using $(X_i, y_i)$ with learning rate $\eta$
            \ENDFOR
            \STATE Evaluate loss $\mathcal{L}_i^{(k)}$ on $(X_i, y_i)$
            \IF{$\mathcal{L}_i^{(k)} < \mathcal{L}^\star$}
                \STATE $f^\star \leftarrow f^{(k)}, \; \mathcal{L}^\star \leftarrow \mathcal{L}_i^{(k)}$
            \ENDIF
        \ENDFOR
        \STATE Add noise: $f^\star \leftarrow f^\star + \delta_i$, \quad $\delta_i \sim \mathcal{N}(0, \sigma_p^2 I)$
        \STATE Append $f^\star$ to $\mathcal{M}$
    \ENDFOR
    \STATE \textbf{Aggregate:} $\theta \leftarrow \frac{1}{N} \sum_{f \in \mathcal{M}} f$
\ENDFOR
\RETURN Final global model $f_\theta$
\end{algorithmic}
\end{algorithm}
\section{Experimental Results}\vspace{-0.2cm}
In order to experimentally validate our proposed federated learning framework, we constructed a synthetic dataset to demonstrate the model's behavior in a controlled setting. Each data sample was composed of a 10-dimensional feature vector, where values were drawn from a uniform distribution over the range $[0, 1]$. The binary labels were assigned using the following nonlinear rule:

\[
y = 
\begin{cases} 
1 & \text{if } \sum_{i=1}^{10} x_i > 5 \\
0 & \text{otherwise}
\end{cases}
\]
The approach of labeling introduces non-trivial decision boundary. To simulate client heterogeneity, we generated independent local datasets using different random seeds across clients.

After generating the synthetic dataset, we evaluated ten mutated versions of the global model. Each mutation represented a locally perturbed and evolved variant of the global model on an individual client. 
\begin{table}[h!]
\centering
\begin{tabular}{|c|c|}
\hline
\textbf{Mutation ID} & \textbf{Accuracy} \\
\hline
M1 & 0.973333 \\
M2 & 0.953333 \\
M3 & 0.963333 \\
M4 & 0.946667 \\
M5 & 0.933333 \\
M6 & 0.943333 \\
M7 & 0.943333 \\
M8 & 0.873333 \\
M9 & 0.946667 \\
M10 & 0.916667 \\
\hline
\end{tabular}
\vspace{0.35 cm}
\caption{Performance of 10 Mutated QENN Models.}
\label{tab:mutation_results}
\end{table}

\vspace{-1.75 cm}
\begin{figure}[h!]
    \centering
    \includegraphics[width=1.15\textwidth]{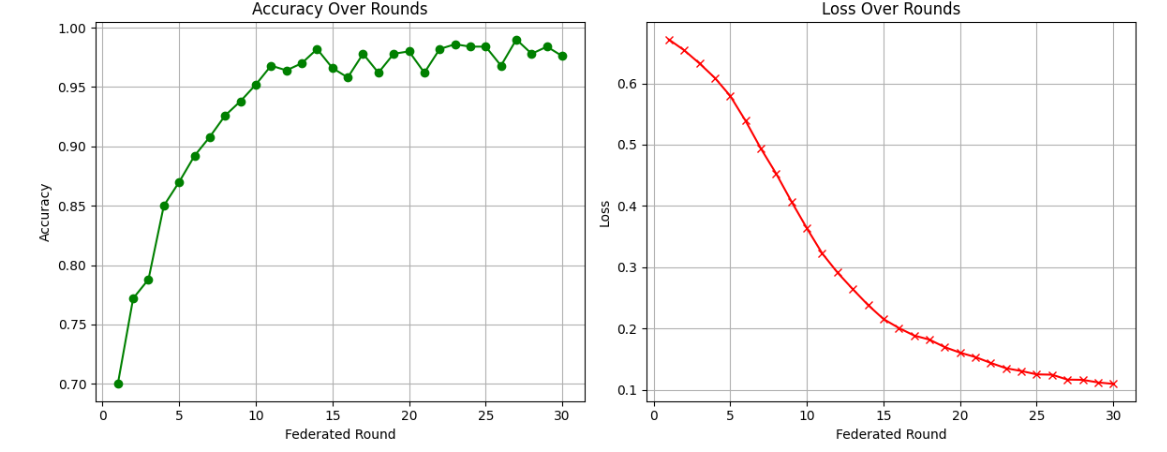}
    \caption{Accuracy of Global Model on Synthetic Dataset over Training Rounds}
    \label{fig:accuracy}
\end{figure}
As shown in Table~\ref{tab:mutation_results}, mutation M1 achieved the highest accuracy at 0.973. Most of the models stayed around the 0.94-0.96 range, with some outliers. This distribution confirms the strength of our mutation-selection mechanism in improving local model quality.

Finally, we tracked the performance of the federated QENN over multiple training rounds on the synthetic dataset. Figure~\ref{fig:accuracy} shows that the accuracy of the global model consistently improves and stabilizes around 0.97, while the loss decreases and plateaus near 0.1. These trends validate the theoretical advantages of our evolutionary approach in a federated learning context.

To evaluate the performance of the model, we compare the performance of the baseline and federated approaches across three datasets: MNIST, CIFAR10, and CIFAR100. The metrics evaluated are accuracy, F1 score, and cross-entropy loss. The results are visualized below.

\begin{figure}[h!]
    \centering
    \begin{subfigure}[b]{0.32\textwidth}
        \centering
        \includegraphics[width=\textwidth]{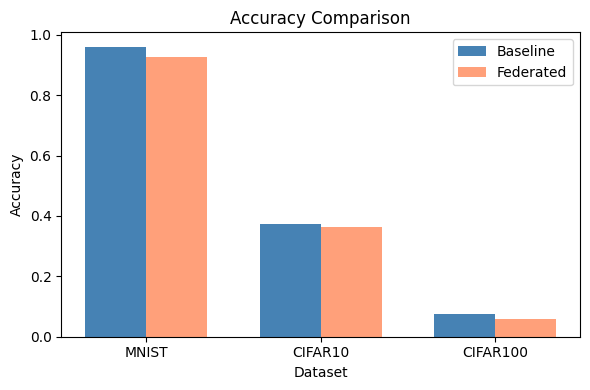}
        \caption{Accuracy comparison}
        \label{fig:accuracy}
    \end{subfigure}
    \hfill
    \begin{subfigure}[b]{0.32\textwidth}
        \centering
        \includegraphics[width=\textwidth]{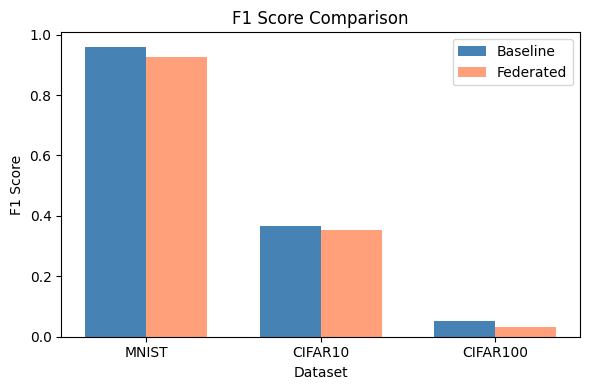}
        \caption{F1 score comparison}
        \label{fig:f1_score}
    \end{subfigure}
    \hfill
    \begin{subfigure}[b]{0.32\textwidth}
        \centering
        \includegraphics[width=\textwidth]{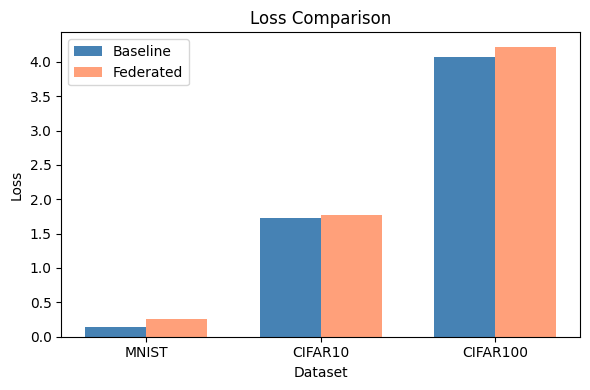}
        \caption{Loss comparison}
        \label{fig:accuracy}
    \end{subfigure}
    \caption{Performance comparison across MNIST, CIFAR10, and CIFAR100 for accuracy, F1 score, and loss.}
    \label{fig:comparison}
\end{figure}

\vspace{-1 cm}

\section{Discussion}

\subsection{Model Limitations}
The model, while theoretically grounded, displayed some key limitations under realistic conditions. The model struggled with highly non-IID data distributions, where each agent only had access to a small biased subset of the overall dataset. Additionally, the model was very sensitive to client dropout. While moderate levels of dropout sometimes helped by filtering out less effective updates, higher dropout rates disrupted training stability and led to a noticeable decline in accuracy.

\subsection{Effectiveness against Datasets}
Reviewing the accuracy of the model’s with the mutations, it is apparent that incorporating mutations returns a wide range of accuracy. Even if some mutated models underperform compared to a few others, some mutations lead to reasonable performance improvements. This difference demonstrates the importance of pursuing a large set of mutated models instead of exploring just one global model.
The small drop in accuracy and F1 score with a slight increase in loss during federated training is attributed to data sharding being done at the client end, resulting in less coherent data updates and diverse data distributions. We focused on smoothing these differences disparities through tuning of local epochs, learning rates, and model aggregation in such a way that the global model could be aligned with the centralized configuration. It is noteworthy that the variations in performance between the federated and the baseline systems were quite close, which indicates that there is some degree of reduced accuracy but still supports the proposition that federated learning is an approach worthy of consideration for those concerned about data privacy, legal restrictions, and dispersed model training.
\vspace{-0.2 cm}
\section{Future Work}
In this paper, we have introduced a framework for Quantum-Evolutionary Federated Learning, combining quantum computing principles, evolutionary algorithms, and federated learning. While the results are promising, several avenues for future work remain. Applying the model to real-world federated datasets would offer deeper insight into its performance under practical constraints. 
\vspace{-0.2 cm}

\bibliography{bibliography}

\end{document}